# A Comparative Study on Machine Learning Algorithms for the Control of a Wall Following Robot

Issam Hammad, Kamal El-Sankary, and Jason Gu

Electrical and Computer Engineering Department, Dalhousie University, Halifax, NS, Canada, B3H 4R2

**For Citation Please Use:**

Issam Hammad, Kamal El-Sankary, and Jason Gu. " A Comparative Study on Machine Learning Algorithms for the Control of a Wall Following Robot." *2019 IEEE International Conference on Robotics and Biomimetics (ROBIO)*. IEEE, 2019. (Pages: 2995 – 3000)

**IEEE URL:** https://ieeexplore.ieee.org/document/8961836

# A Comparative Study on Machine Learning Algorithms for the Control of a Wall Following Robot

Issam Hammad, IEEE Student Member, Kamal El-Sankary, IEEE Member, and
Jason Gu, IEEE Senior Member

Electrical and Computer Engineering Department, Dalhousie University, Halifax, NS, Canada, B3H 4R2
Email: issam.hammad@dal.ca

*Abstract*—A comparison of the performance of various machine learning models to predict the direction of a wall following robot is presented in this paper. The models were trained using an open-source dataset that contains 24 ultrasound sensors readings and the corresponding direction for each sample. This dataset was captured using SCITOS G5 mobile robot by placing the sensors on the robot waist. In addition to the full format with 24 sensors per record, the dataset has two simplified formats with 4 and 2 input sensor readings per record. Several control models were proposed previously for this dataset using all three dataset formats. In this paper, two primary research contributions are presented. First, presenting machine learning models with accuracies higher than all previously proposed models for this dataset using all three formats. A perfect solution for the 4 and 2 inputs sensors formats is presented using Decision Tree Classifier by achieving a mean accuracy of 100%. On the other hand, a mean accuracy of 99.82% was achieves using the 24 sensor inputs by employing the Gradient Boost Classifier. Second, presenting a comparative study on the performance of different machine learning and deep learning algorithms on this dataset. Therefore, providing an overall insight on the performance of these algorithms for similar sensor fusion problems. All the models in this paper were evaluated using Monte-Carlo cross-validation.

I. INTRODUCTION

Fully autonomous mobile robots are used in various industries today such as nuclear power plant, oil refineries, chemical factories, and military applications. In general, autonomous mobile robots play an important role in process control applications. One of the most critical tasks that these robots should accomplish is navigation by following a wall. Wall following can be used in several operations such as fault detection, search and rescue and in detections of cracks in oil pipelines [1]. Achieving a highly accurate control for these robots is vital for their intended operations. Any small error in the accuracy could be costly and it could result in missing the inspection of a certain portion of the wall or pipe. Several papers have proposed different control methods for the wall following algorithms such as in [1-3]. As an example, the design in [3] uses a data-driven fuzzy controller learned through differential evolution to control a hexapod robot. The design in [2] measured the readings of 24 ultrasound sensors and recorded the corresponding direction to perform a wall following task. This resulted in having a problem which is non-linearly separable. Therefore, [2] has proposed several neural network designs including Multi-Layer Perceptron (MLP) and Elman Recurrent Network. The authors of [2] have published the dataset for the 24 sensors readings and the corresponding direction that the robot should follow in the repository [4]. Several other following research papers have proposed different designs for the controller based on the dataset in [4] such as the designs in [5-10].

With the recent advancements in machine learning, the state-of-art of many applications has seen a significant advancement. Machine Learning is used today in applications such as image recognition and classification, natural language processing, self-driving cars, healthcare, and financial fraud detection. Machine learning can also be utilized to build a highly accurate controller for a mobile robot as will be presented in this paper. Decision Tree (DT) is one of the most popular and straight forward methods of machine learning. DT for certain problems can be improved by employing ensemble learning and boosting techniques. Random Forest Classifier (RFC) and Gradient Boosting Classifier (GBC) are examples of that.

Deep learning [11] is another subset of machine learning that was proven to be one of the most powerful methods nowadays especially for classification problems with large datasets. Deep learning allows for the training of deep neural networks which are composed of multiple hidden layers. While the principle of training neural networks with multiple hidden layers is relatively old, the lack of computational power and available data in the past have imposed a major challenge on the advancement of this field [12]. The available computational power and data available nowadays allow for the implementation of deep complex neural networks.

One of the challenges in machine learning is to select the right algorithm for the intended problem. According to the popular No Free Lunch Theorem [13], there is no golden machine learning algorithm that can outperform all the other machine learning algorithms in solving all possible problems. This paper employs the sensor fusion problem in [4] to evaluate and compare the accuracies of the most popular machine learning algorithms for this problem and similar problems. Besides comparing the performance of different models to solve for the problem in [4], this paper aims to provide a research insight to solve other data fusion problems with similar data characteristics.

This paper is divided as follows, section II provides detailed information about the dataset in [9] and the previously proposed designs. Section III demonstrates the performance of different proposed machine learning and deep learning models. In section IV a comparison between this paper's models and previously proposed models is illustrated. Section V summarizes the research conclusion and findings.

## II. DATASET DETAILS AND LITERATURE REVIEW

The dataset [4] was constructed using SCITOS G5 mobile robot by applying a wall following algorithm and then collecting the ultrasound sensors readings. Fig. 1 shows the SCITOS G5 robot [14] which was used for data collection. SCITOS G5 is a mobile robot platform for research and industrial applications [15]. A total of 24 sensors were installed on the robot's waist and their recordings were collected. Each record in the dataset consists of the 24 ultrasound sensors readings and the corresponding direction that the robot should follow. These directions are divided into the following 4 classes: move forward, slight right-turn, sharp right-turn, and slight left-turn. The data was collected at a rate of 9 samples per second [2]. A total of 5456 samples were captured.

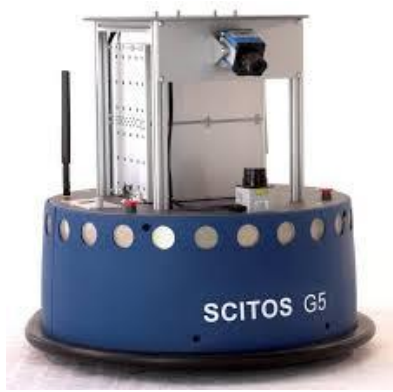

Fig. 1. SCITOS G5 mobile robot

The 24 sensors were placed 15° apart from each other. Each sensor had different minimum, maximum, mean and standard deviation value. The detailed statistical information for the dataset can be found in [4]. In addition to the 24 full sensor readings, [4] has provided simplified formats of this dataset. One with 4 sensor readings per sample and the other with 2 sensor readings per sample. The 4 sensors dataset was constructed from the full 24 sensors version by taking the minimum sensor reading within a 60-degree arc for each direction, the front, the left, the right and the back of the robot. The 2 sensors dataset is even a more simplified version constructed by taking only the left and the front sensor readings from the 4 sensors dataset.

Several models were proposed for a robot controller based on the dataset in [4] such as the designs in [5-10]. These designs used different techniques for building the controller which are presented in section IV. Some of the reported accuracies in these designs were based on the training accuracy not a separate test set accuracy. The training accuracy doesn't provide an accurate metric on how the system can generalize to new data, therefore, these accuracies can be disputed. According to [16] the data for the training the model shouldn't be used for testing. A separate test set should be reserved for the purpose of performance evaluation. A known problem that occurs when the model is perfectly fitting on the training data is the problem of data overfitting. When overfitting occurs, the accuracy on the training set is usually high, however, the model performs poorly on the test set or any new data. In this paper, the overall accuracy will be assessed using a separate test set as will be explained in the next section.

## III. MODELS COMPARASION AND EVALUATION

This section presents the testing results for several machine learning algorithm using the dataset [4]. As mentioned in the previous section, the dataset [4] provides three different formats. The full one 24 sensor inputs, while the other two are simplified to 4 and 2 sensor inputs. Fig. 2 depicts the various machine learning and deep learning algorithms that are evaluated in this section using all three dataset formats. While deep learning is a subset of machine learning, it is common nowadays to separate deep learning from the rest of the machine learning algorithms for comparison purposes. The popular machine learning python libraries Keras [17] and Scikit-learn [18] were used to implement the models in this paper. All models were executed using Google Colaboratory machine learning tool. To demonstrate the true performance of the proposed models, Monte-Carlo cross-validation was applied. Fig. 3 depicts the applied steps for model evaluation using Monte-Carlo cross-validation [19]. As can be seen from the figure, the first step is to shuffle the data randomly, then the data is split into a separate training set and test set. In this experiment, a ratio of 10:1 was used for the splitting, this resulted in having a 4910 training sample and 546 testing sample. The steps in Fig. 3 are repeated for n iteration, then the mean of accuracy for all iterations is obtained. In this paper, the value of n=50 iterations was used to evaluate all the proposed models.

Several machine learning models were tested and evaluated. This includes Decision Tree (DT), Gradient Boost Classifier (GBC), Random Forest Classifier (RFC), Linear Discriminant Analysis (LDA), Support Vector Machine (SVM), K-Nearest

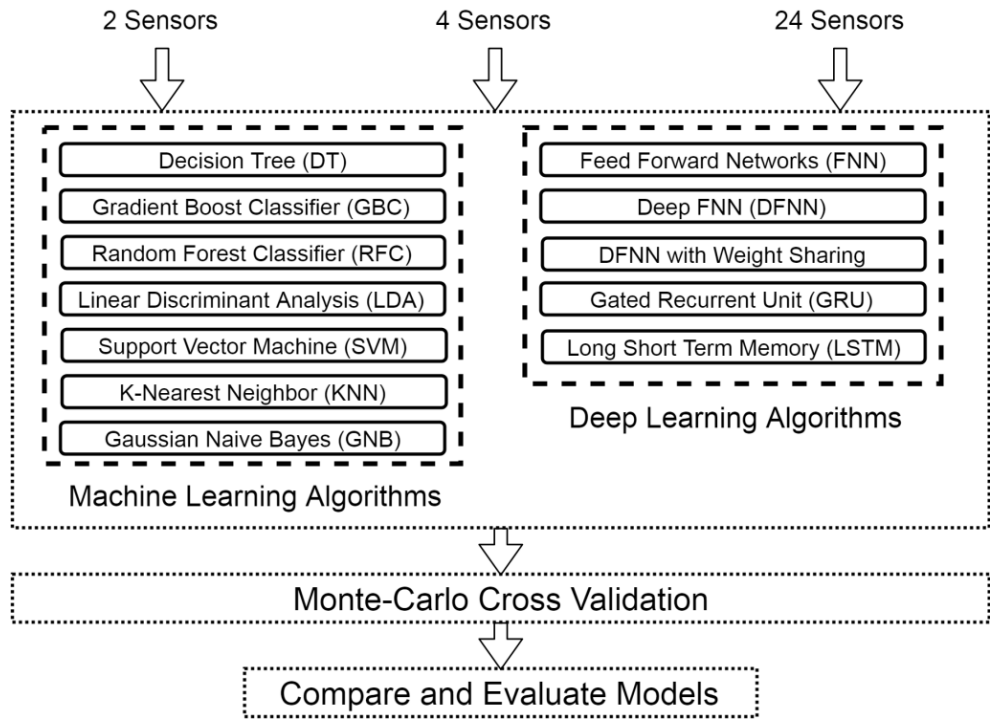

Fig 2. Evaluated machine learning and deep learning models

Neighbour (KNN). Table I, demonstrates the reported mean accuracy for all the tested machine learning and deep learning models for all three sensor inputs configurations. As can be seen in the table, a perfect solution with a mean accuracy of 100% is presented for the simplified dataset with 4 or 2 input sensors using DT.

Unlike previous models which proposed more complicated solutions for the simplified dataset, a simple Decision Tree model outperform all the previously proposed models in terms of accuracy. Fig. 4 depicts the DT model that achieves this perfect accuracy. In the figure, $X_0$ is a notation for the front sensor while $X_1$ is a notation for the left sensor.

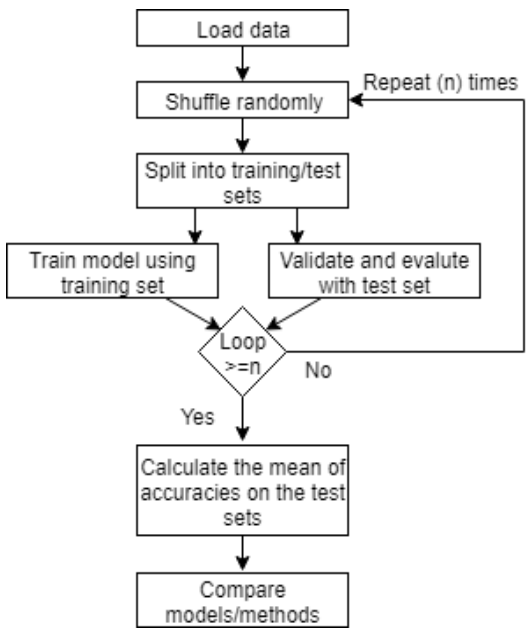

Fig. 3. Model evaluation using Monte-Carlo cross-validation

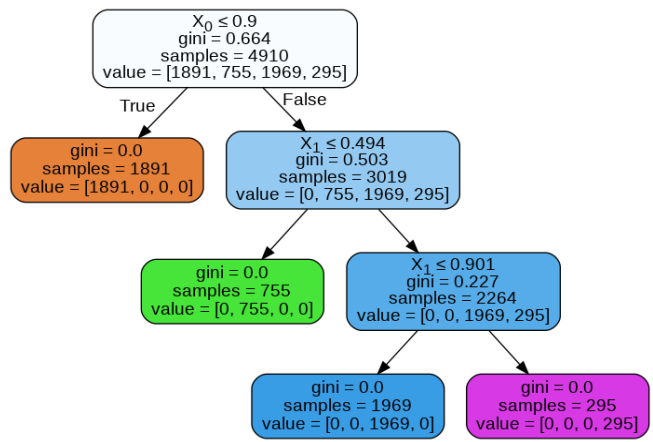

Fig 4. DT perfect solution for the simplified dataset

The solution for both the 4 sensor inputs dataset and the 2 sensor input dataset is identical using DT as only the front and left sensors are used in the tree in both cases. Other algorithms such as GBC and RFC achieves almost a perfect accuracy using the simplified dataset, while others performed very poorly such as LDA. For the full dataset using 24 sensors. The GBC outperform every other machine learning model with a mean accuracy of 99.82%. This is achieved accuracy outperform all previously proposed models as well. Using an implementation with the simplified version can have a disadvantage in terms of possible sensor failure or noise, therefore achieving a high accuracy using all 24 sensor inputs carries a value from that perspective. While the performance of the presented machine learning models is very promising, it is worthwhile to explore how the deep learning models perform on this dataset as well. This provides an intuition for the performance of machine learning vs. deep learning for problems with similar dataset characteristics. Deep learning is very powerful for large datasets and primarily for classification problem. The dataset [4] presents a classification problem as it contains 4 possible output actions. Deep learning uses deep neural networks in the form of Feedforward Networks (FNN), Convolutional Neural Networks (CNN), Recurrent Neural Networks (RNN), and others. As part of this study, several FNNs and RNNs networks were implemented and tested. The RNN networks included Gated Recurrent Units (GRUs) and Long Short-Term Memory (LSTM) implementations. The deep learning architecture that achieved the highest accuracy is depicted in Fig. 5 which is a Deep FNN (DFNN) model with a weight sharing technique. This architecture outperforms all previously proposed FNN solutions with a mean accuracy of 98.1%. As can be seen in Fig. 5, this architecture has an input layer of 24 neurons, where each neuron has a vector input of the entire 24 sensors. Each neuron in this layer assigns one shared weight and bias per vector input. This layer is different than the regular FNN layer in which every neuron process one singular value only. The output of this layer is a 2D matrix with a size of (24,24), or a 3D matrix if we take the batch size into account with a size of (batch_size,24,24). Following the input layer, the 2D output is unrolled and is connected to three feed-forward hidden layers. The hidden layers are of sizes 16, 8 and 4 respectively. The output layer has 4 units with a softmax activation function. The softmax activation function will provide a probabilistic value for each action. The action with the highest probability will be followed by the robot. In order to avoid overfitting, a dropout regularization of 10% was applied in each layer of the network. Additionally, the input data was normalized, and the output of each layer was also normalized, this is known as batch normalization. This model was trained with a batch size of 32 and using 200 Epochs. The model used Adadelta [20] optimizer and Categorical Crossentropy loss function. By comparing the accuracies achieved by the machine learning and the deep learning models, it can be clearly seen that the machine learning models are more suitable to use for this sensor fusion problem.

TABLE I. MACHINE LEARNING AND DEEP LEARNING MODELS ACCURACY

| Deep Learning Models | | | |
|---|---|---|---|
| Model | Mean Accuracy (24 Sensors) | Mean Accuracy (4 Sensors) | Mean Accuracy (2 Sensors) |
| DFNN with Weight Sharing | 98.1% | 96.8% | 95.7% |
| DFNN (3 Hidden Layers) | 96.4% | 92.5% | 90.6% |
| FNN (1 Hidden Layer) | 94.14% | 90.1% | 88.3% |
| Gated Recurrent Unit (GRU) | 94.69% | 96.52% | 95.05% |
| Long Short Term Memory (LSTM) | 94.13% | 96.15% | 94.87% |
| Machine Learning Models | | | |
| Model | Mean Accuracy (24 Sensors) | Mean Accuracy (4 Sensors) | Mean Accuracy (2 Sensors) |
| Decision Tree (DT) | 99.52% | 100% | 100% |
| Gradient Boost Classifier (GBC) | 99.82% | 99.94% | 99.96% |
| Random Forest Classifier (RFC) | 99.42% | 99.93% | 99.97% |
| Linear Discriminant Analysis (LDA) | 65.85% | 71.31% | 70.65% |
| Support Vector Machine (SVM) | 90.36% | 92.60% | 93.75% |
| K-Nearest Neighbour (KNN) | 86.83% | 96.45% | 98.43% |
| Gaussian Naïve Bayes (GNB) | 52.78% | 89.10% | 90.61% |

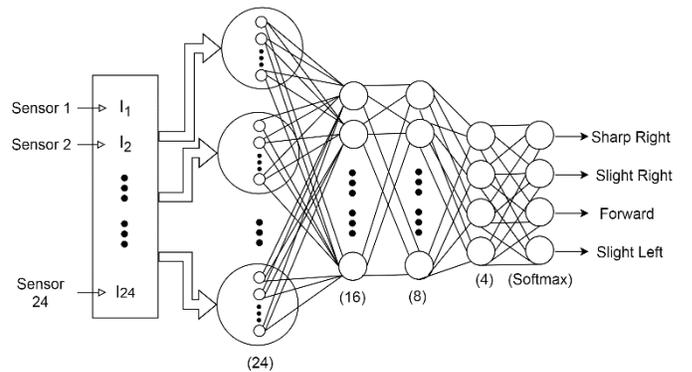

Fig. 5. Deep learning model with weight sharing

IV. COMPARASION WITH PREVIOUS DESIGNS

This section presents a comparison between the proposed models in this paper and the models proposed in previous papers. Table II demonstrates a comparison with different previous designs that were implemented by employing the dataset [4]. The table is divided into three main sections based on the number of input sensors used in the models. As can be seen from the table the models proposed in this paper outperform all previous models in terms of accuracy. Additionally, the table indicates the models that applied cross-validation by splitting the dataset into a separate training set and a test set. Models that didn't apply this split often report the training accuracy as the model accuracy. As mentioned earlier, this doesn't provide an accurate evaluation of how the model can generalize to new data. The DT model that was proposed in the previous section of this paper provides a perfect solution with a mean accuracy of 100% for the 2 and 4 sensors datasets. This perfect accuracy wasn't achieved by any previous model. Additionally, using GBC for the 24 sensors dataset achieves a mean accuracy of 99.82% which is also the highest compared to all previous designs.

TABLE II. COMPARISON WITH PREVIOUSLY PROPOSED MODELS

| 2 Sensors Dataset | | | |
|---|---|---|---|
| **Paper** | **Model Description** | **Accuracy** | **Train/Test Split** |
| This | Decision Tree (DT) | 100% | ✓ |
| [7] | Particle swarm optimization | 98.8% | ✓ |
| [2] | Multi Layer Perceptron (Neural Network) | 97.59 % | ✗ |
| [5] | Shallow Neural Network | 92.67% | ✗ |
| [2] | Elman Recurrent | 96.42% | ✗ |
| **4 Sensors Dataset** | | | |
| **Paper** | **Model Description** | **Accuracy** | **Train/Test Split** |
| This | Decision Tree (DT) | 100% | ✓ |
| [8] | Bayesian Network | 93.3% | ✓ |
| [9] | Adaptive Resonance Theory-1 | 86.69% | ✓ |
| [5] | Shallow Neural Network | 81.32% | ✗ |
| **Model 24 Sensors Dataset** | | | |
| **Paper** | **Description** | **Accuracy** | **Train/Test Split** |
| This | Gradient Boost Classifer (GBC) | 99.82% | ✓ |
| [10] | Probablistic Neural Network | 99.63% | ✓ |
| [9] | Adaptive Resonance Theory-1 | 99.59% | ✓ |
| [7] | Particle swarm optimization | < 80% | ✓ |
| [6] | Shallow Neural Network | 69.72% | ✗ |

## V. CONCLUSION

This paper presented a comparison of the performance of various machine learning and deep learning models for a wall following robot controller. An open-source dataset containing ultrasonic sensors readings and the corresponding robot direction was employed in this study. The models were trained using three formats of the datasets, a full format with 24 sensor inputs and two simplified ones with 4 and 2 sensor inputs. The most popular machine learning and deep learning algorithms were implemented and evaluated. A perfect solution using a DT model was proposed for the simplified dataset which achieves a mean accuracy of 100%. On the other hand, an accuracy of 99.82% was achieved for the full dataset with 24 input sensors using GBC. The proposed models in this paper outperform all the previously proposed models for this dataset in terms of accuracy. All models were evaluated using Monte-Carlo cross-validation.

## VI. ACKNOWLEDGEMENT

This work was supported by The Killam Trusts Scholarship.